\documentclass[10pt,twocolumn]{IEEEtran} 

\def\BibTeX{{\rm B\kern-.05em{\sc i\kern-.025em b}\kern-.08em
    T\kern-.1667em\lower.7ex\hbox{E}\kern-.125emX}}

\title{Memristor-based Deep Convolution Neural Network: A Case Study}

\author{\large Fan Zhang, Miao Hu \\ 
Binghamton University, Binghamton, NY USA\\
E-mail: fzhang27@binghamton.edu, miaohu@binghamton.edu}

\usepackage{xfrac}  
\usepackage{graphicx}
\usepackage{subcaption}
\graphicspath{ {images/} }
 \newcommand\quotient[2]{
        \mathchoice
            {
                \text{\raise1ex\hbox{$#1$}\Big/\lower1ex\hbox{$#2$}}%
            }
            {
                #1\,/\,#2
            }
            {
                #1\,/\,#2
            }
            {
                #1\,/\,#2
            }
    }

\usepackage{comment}
\usepackage[usenames,dvipsnames]{xcolor}
\usepackage{amsmath}
\usepackage{algorithm}
\usepackage[noend]{algpseudocode}
\include{localdefs}
\usepackage{color}
\usepackage{etoolbox}
\newbool{inccomment}
\setbool{inccomment}{true}
\newcommand{\zf}[1]{\ifbool{inccomment}{{\color{magenta}#1}}{}}
\newcommand{\hm}[1]{\ifbool{inccomment}{{\color{blue}#1}}{}}

\newcommand\Tstrut{\rule{0pt}{2.6ex}}       
\newcommand\Bstrut{\rule[-0.9ex]{0pt}{0pt}} 
\newcommand{\TBstrut}{\Tstrut\Bstrut} 

\makeatletter
\def\BState{\State\hskip-\ALG@thistlm}
\makeatother
\algnewcommand{\parState}[1]{\State%
  \parbox[t]{\dimexpr\linewidth-\algmargin}{\strut #1\strut}}
  
\begin{document}

\maketitle

\begin{abstract}
In this paper, we firstly introduce a method to efficiently implement large-scale high-dimensional convolution with realistic memristor-based circuit components.
An experiment verified simulator is adapted for accurate prediction of analog crossbar behavior.
An improved conversion algorithm is developed to convert convolution kernels to memristor-based circuits, which minimizes the error with consideration of the data and kernel patterns in CNNs.
With circuit simulation for all convolution layers in ResNet-20, we found that 8-bit ADC/DAC is necessary to preserve software level classification accuracy.  
\end{abstract}


\section{\bf Introduction}

Convolutional Neural Networks(CNN) have led to many performance breakthroughs in image classification, video object tracking, and audio processing applications\cite{lecun_deep_2015}. 
Since AlexNet won the ILSVRC 2012 and is more than 10\% lower than other contemporary competitors on top-5 error rate\cite{krizhevsky_imagenet_2012}, CNN evolved into many different models, such as GoogLeNet\cite{szegedy_going_2015}, VGG\cite{simonyan_very_2014}, ResNet\cite{he_deep_2015}.
Using Graphics Processing Unit(GPU) to accelerate convolution computations plays an important role in CNN's success, since convolution is the most time-consuming part in CNN, and the throughput of convolution is prior than accuracy\cite{gupta_deep_2015}. 
However, for large scale Convolutional Neural Networks(CNN), modern computing hardware often encounter on-chip memory bottleneck when dealing with high volume convolution calculations\cite{chandrasekhar_compression_2017,rhu_vdnn:_2016}.

Recently, many researchers show interest on emerging memristor crossbar for its computing-in-memory feature\cite{gao_demonstration_2016,adam_3-d_2017,hu_memristor-based_2018,liu_harmonica:_2016,yu_neuro-inspired_2018,li_analogue_2018}. 
A memristor crossbar can carry out large-scale analog vector matrix multiplication in one step by collecting analog output with input signal flowing through the array, where weights are stored non-volatilely at cross-point memristor devices. 
Since weight storage and weighted summation both happen at the same location of memristors, it enables ultra-high computing efficiency as well as low communication requirement.  
Overall, memrisotr crossbar-based implementation may overcome von-Neumann architecture's shortage and are suitable for deep learning inference functions.

However, integrating analog memristor crossbars with digital circuits is not trivial. One of the key issues is that a rigorous circuit simulation is still missing for memristor-based CNN implementations, especially on the modeling of analog memristor crossbar behavior.
As SPICE is too slow on simulating large-scale memristor crossbar arrays, specific tools have been developed to investigate memristor-based neural network (NN) implementations\cite{xia_mnsim:_2016}\cite{chen_neurosim:_2018}.
Although they provide accurate estimation for speed, area and power, the prediction of functionality is questionable due to over-simplified memristor crossbar models, which is also lack of experiment verification at large scale.
It leads to a serious question: \textit{if we implement state-of-the-art CNNs with memristors, will it really work?}
This question can only be addressed with an analog circuit simulation for all the memristor crossbars in CNNs, since it is the most immature part in the whole system. 

In this paper, we investigate memristor-based large-scale CNN implementation with experiment verified memristor crossbar simulators. Our contributions are listed as below: 
\begin{itemize}
\item First, we provided an efficient mapping method to map high dimension kernels to 2-D memristor crossbars using realistic circuit components. 
\item Second, we developed an improved conversion algorithm to convert kernel to crossbar conductance with consideration of non-ideal hardware as well as data/kernel patterns. 
\item Third, we addressed the aforementioned question with ResNet-20 on cifar-10. A careful simulation is done at all convolution layers to capture the error propagation in CNN. 
Our result shows that 8-bit ADC/DAC quantization is necessary to preserve the classification accuracy.  
\end{itemize}


\section{\bf Priliminary}

\subsection{\bf Deep Residual Neural Network}


Deep residual neural network(ResNet) was firstly introduced in ILSVRC 2015 \cite{he_deep_2015}. 
So far, ResNet is the state-of-the-art CNN on image classification problems. 
An ensemble of residual nets up to 152 layers achieves 3.57\% error on the ImageNet test set and won the 1st place on the ILSVRC 2015 classification competition. 
To evaluate how state-of-the-art CNNs perform on memristor crossbar arrays, ResNet is an ideal case study. 

\begin{figure}[t]
      \centering
      \includegraphics[width=0.4\textwidth]{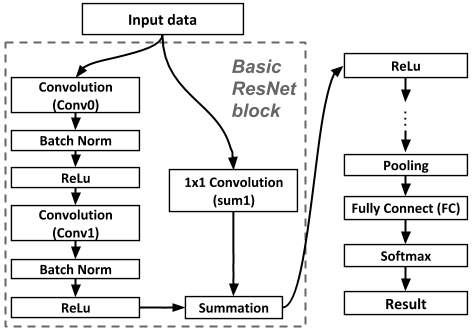}
      \caption{Basic block diagram of ResNet.}
      \label{fig_resnet}
\end{figure}

Fig.\ref{fig_resnet} shows the basic block diagram of ResNet. It combines multiple  convolution, batch normalization, and rectified linear units(ReLU) together as its basic building block.   
Different from other CNNs, ResNet uses a shortcut to add input data to the output result of a block.   
If two data inputs have different dimensions at the summation stage, a $1 \times 1$ convolution layer will be introduced in shortcut to match the dimension.
The summation result is sent to ReLU and pass to the next Block. 
At the end of ResNet, pooling layer, one or more Fully Connected (FC) layers, and a softmax layer are used in sequence to generate final classification result.  

\subsection{\bf 1T1M crossbar for VMM}

\begin{figure}[t]
      \centering
      \includegraphics[width=0.45\textwidth]{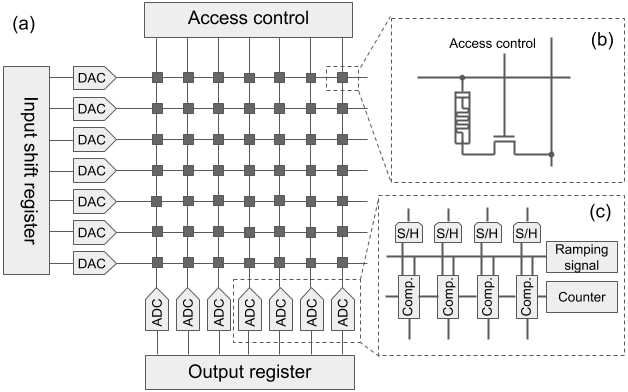}
      \caption{1T1M crossbar for VMM operation. (a) shows the overall circuit diagram; (b) shows the 1T1M cell (c) shows the ramping ADC design.}
      \label{fig_xbar}
\end{figure}

Fig.\ref{fig_xbar}(a) shows the general structure of a one-transistor-one-memristor (1T1M) crossbar for vector matrix multiplication (VMM).
In this structure, memristor crossbar is the core component as it enables analog current weighted summation, which leads to ultra efficient VMM operation. 
A $m \times n$ memristor crossbar is made by $m$ row and $n$ column metal wires where memristors formed at intersecting points. 
Each memristor can be tuned to arbitrary conductance within its programmable range.
To enable precise, non-disturbing tuning in large crossbar arrays, 1T1M cell (Fig.\ref{fig_xbar}(b).) is necessary to program the entire array with arbitrary conductance matrix $G$.
By applying voltage inputs $V$ at the rows simultaneously and read current outputs $I$ from the columns, analog weighted summation is achieved through Kirchhoff's Current Law and Ohm's Law.
In an ideal crossbar, input-output relationship can be represented as below:
$$
I = VG
$$
By mapping input vector $X$ to input voltage $V$, matrix $A$ to conductance $G$, and output current $I$ back to output result $Y$, a memristor crossbar can be regarded as an analog VMM module to realize $Y=XA$ in one step. Note that $I = VG$ is only valid for ideal crossbar while wire resistance can be ignored and device conductance is independent of voltage/current. In real crossbars, the input-output relationship is far more complex and needs analog circuit simulation. 


Fig.\ref{fig_xbar}(c) illustrates a ramping ADC design\cite{delagnes_low_2007}, it uses a shared ramping signal generator and counter to support multiple channels simultaneously. The ramping signal generator produces increasing/decreasing analog voltage signal to all comparators, and its analog value is synchronized with the digital representation in the counter. Comparators compare the crossbar output to the ramping signal. Once a flip is detected, comparator captures the current value in the counter, which is the corresponding digital representation of crossbar output. This concept can be applied to DAC design as well. Overall, The ramping ADC/DAC design can achieve lower area and power for multi-channel applications, and it is a suitable digital interface for crossbar-based accelerators.   



\subsection{\bf Experiment-Verified Crossbar Simulator and The Conversion Algorithm}
So far, simulation tools like MNSIM and NeuroSim provide excellent architecture-level analysis.
However, their memristor crossbar models are over simplified. 
In MNSIM\cite{xia_mnsim:_2016}, an estimated behavior model is used instead of solving node equations; In NeuroSim\cite{chen_neurosim:_2018}, the impact of wire resistance is simplified by adding wire resistance to cross-point devices.
All of these over-simplifications result in unignorable error between predicted output to real output of crossbars. 
And this error become even more severe in large scale implementations. 
In short, coarse modeling of crossbar may be acceptable on power/area/speed estimations, but is questionable on functionality evaluations, such as computing accuracy at layers, or classification accuracy of NN implementations.  
After all, to evaluate NN's functionality on memristor-based implementations, an accurate crossbar simulator is necessary.  

\begin{figure}[t]
      \centering
      \includegraphics[width=0.4\textwidth]{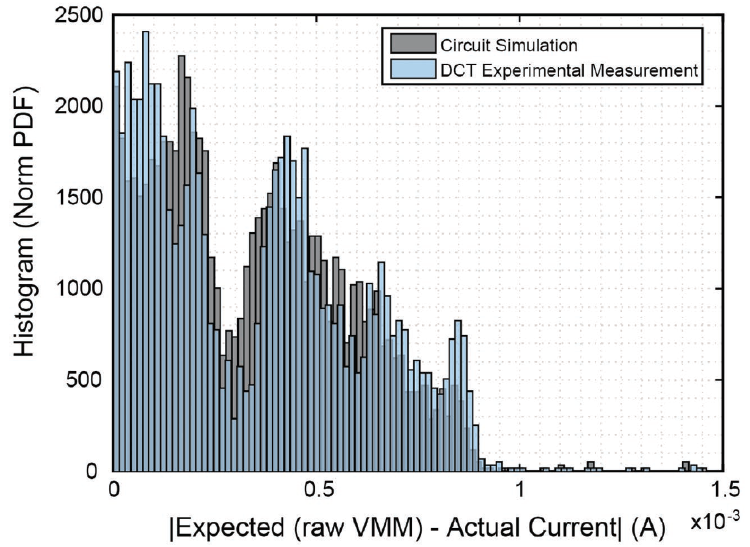}
      \caption{1T1M simulation and experiment result \protect\cite{hu_memristor-based_2018}. Wire segment resistance is calibrated to 1$\Omega$, transistor and memristor models are calibrated with in-array device test.}
      \label{fig_1T1Msim}
   \end{figure}

An accurate crossbar simulator should use realistic device models, solve node equations with consideration of wire resistance and other circuit parameters, then finally verified with experiment result at large scales.
Here We adapt the experiment-verified memristor crossbar simulator from \cite{hu_memristor-based_2018}.
Fig.\ref{fig_1T1Msim} illustrates the simulation result of the memristor crossbar simulator, which is verified with experiment up to $128 \times 64$.

Besides the simulator, the conversion algorithm, as a mapping method is employed to improve VMM computing accuracy with consideration of wire resistance and device nonlinearity\cite{hu_dot-product_2016}.
The conversion algorithm uses a conversion signal $V_{conv}$ to find a new conductance mapping $G'$, so that
$$
VG \approx \mathrm{CrossbarSim}(V, G')
$$
where CrossbarSim() is the realistic crossbar model. 
Although the original conversion algorithm provides 7$\sim$8 bit-accuracy result for general VMM, it does not consider impact of data/weight patterns in specific applications, which may affect its performance significantly. 

\section{\bf Methodology}

In this work, we take ResNet-20 as our case study for memristor-based CNN implementation. 
We first provide a dense mapping method to map high dimension kernels onto 2-D memristor crossbar arrays efficiently. 
By doing a near full circuit simulation of memristor-based implementation for ResNet-20, we explain how to optimize the conversion signal for each layer based on its corresponding convolution kernel and data pattern. 
Moreover, we introduce an additional calibration step to further improve the performance and robustness of the circuit.  
Our memristor-based CNN implementation can give precise convolution result for each layer, and its precision is independent of data sparsity and kernel types.

\subsection{\bf Dense mapping for crossbar-based convolution}

In ResNet, each convolution layer has a 4-D kernel, as illustrated in Table.\ref{Table_mapping}. For example, the 3*3*3*16 kernel of 'Conv0' means it has 16 sets of 3*3*3 3-D kernels, and each 3-D kernels contains 3 2-D kernels for the 3 channels (RGB) of color images. 
To perform convolution on memristor crossbar, we need convert high dimensional convolution into 2-D VMM. 
It is well known that 2D convolution can be implemented using matrix multiplication by converting kernel to a Toeplitz matrix. 
Topelitz matrices are sparse matrices which contains many zeros, so we named this method as \textit{sparse mapping}. 
\begin{figure}[t]
      \centering
      \includegraphics[width=0.465\textwidth]{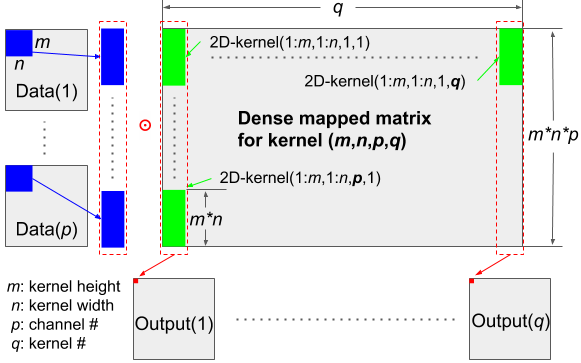}
      \caption{Dense mapping for input data size (j,k,p), where j,k,p means data height, width, and channel, respectively. The circuit needs j*k iterations to process the input data.}
      \label{Fig_densemap}
\end{figure}
   
However, sparse mapping has three major issues:
First, memristor crossbar is by nature a dense array structure with positive values,  not efficient for sparse matrix implementation.
Mapping a sparse matrix to crossbar means that the memristors assigned with zero entries would do nothing but adding error to the final result due to leakage current, as well as waste circuit area.    
Second, sparse mapping requires a huge crossbar array, which is not always feasible, and vulnerable to noise and defects.  
third, sparse mapping requires a large peripheral circuit to support the huge array. 
Since peripheral circuit dominates the total power/area, the hardware implementation of sparse mapping would be too costly to afford.  
%
%
%
%

Fig.\ref{Fig_densemap} illustrates the concept of dense mapping.
In contrast to sparse mapping, we developed a new method, named as \textit{dense mapping}, targeting on using small and dense representations to implement convolutions on memristor crossbars efficiently.   
Each 2-D kernel is unrolled to a vector and mapped to the column of crossbar so that one crossbar can implement multiple 2-D kernels as long as it has enough columns. 
For input signal, only data within the convolution window is converted to the row inputs of crossbar arrays. 
For data with multiple channels, 2-D kernels for different channels are stacked into the same column, as input from different channels can be supplied to different rows and weighted summed together.
An input shift register stores the input data within convolution window at the current iteration, and updates its storage as window moves through the entire data space.
The convolution result for data within the convolution window are collected at the column outputs of crossbar.
In this way, a single memristor-based convolution kernel needs $j$*$k$ iterations for input data with size ($j$, $k$, $p$) where $j$, $k$, $p$ are data height, width, and channel, respectively.  

   

Comparing dense mapping to sparse mapping, it is a trade-off between time multiplexing and space multiplexing.
Sparse mapping uses much more extra hardware to produce result without iteration.
However, its efficiency exponentially drops as data/kernel scale up, which means, more and more devices in a rectangular crossbar are unused for increasing data/kernel size.  

Dense mapping is a more adequate and practical method comparing to sparse mapping. 
It not only achieves 100\% usage of devices, but also easy to implement and provide sufficient performance in speed for CNN applications. 
From Table \ref{Table_mapping}, one classification in ResNet-20 needs 9089 iterations in sequential, if no parallel copies of hardware are used. Note that summation (sum\#) is in parallel of convolutions so it's not counted in total iterations.
Assuming memristor crossbar runs at 100 MHz \cite{hu_dot-product_2016}, for each classification the convolution part takes only 0.09 ms, which is way fast enough for real-time classification requirement. 

\begin{table}[t]
\caption{Dense mapping of ResNet-20 convolution layers for $32 \times 32$ color image classification from cifar-10 dataset}
\centering
\begin{tabular}{|p{1.8cm}|p{1.8cm}|p{1.5cm}|p{2cm}|}
\hline
\textbf{Layer name} & \textbf{Kernel size} & \textbf{Crossbar Size} & \textbf{Iteration per. class.}\TBstrut\\ \hline
Conv0      & 3*3*3*16    & 27*16 & 1024        \TBstrut\\ \hline
Conv1-2      & 3*3*16*16   & 144*16 & 1024       \TBstrut\\ \hline
Sum1       & 1*1*16*16   & 16*16 & 1024       \TBstrut\\ \hline
Conv3-6      & 3*3*16*16   & 144*16 & 1024       \TBstrut\\ \hline
Sum2       & 1*1*16*32   & 16*32 & 256        \TBstrut\\ \hline
Conv7      & 3*3*16*32   & 144*32 & 256       \TBstrut\\ \hline
Conv8-12      & 3*3*32*32   & 288*32 & 256       \TBstrut\\ \hline
Sum3       & 1*1*32*64   & 32*64 & 64        \TBstrut\\ \hline
Conv13     & 3*3*32*64   & 288*64 & 64      \TBstrut\\ \hline
Conv14-18     & 3*3*64*64   & 576*64 & 64       \TBstrut\\ \hline
FC         & 1*1*64*10   & 64*10 & 1        \TBstrut\\ \hline
\end{tabular}
\label{Table_mapping}
\end{table}

\subsection{\bf Improved conversion algorithm}

Algorithm \ref{alg_1} summarizes the flow of crossbar-based convolution with the improved conversion algorithm. Table \ref{Table_functions} explains the important functions in the algorithm. After initialization, if the kernel is already mapped and converted onto crossbars, it will directly jump to the computing step to simulate crossbar-based convolution. So to implement ResNet-20 with multiple convolution layers, we only need to do mapping/conversion once for inferencing.   

\begin{algorithm}
\caption{Crossbar-based convolution with improved conversion algorithm}\label{euclid}
\begin{algorithmic}[1]
\State Initialization: setup crossbar parameters(wire resistance, memristor conductance range, etc...)
\State Get a batch of input data
\State If kernel $K$ is mapped and converted, jump to {\bf computing}.
\State Dense mapping kernel $K$ to conductance matrix $G$
\State Optimize conversion signal
\State $InputVectors \gets \text{Partition input data}$
\State $CaliSample \gets \text{Random pick from } InputVectors$
\State $G' \gets \text{Conversion}(G,V_{conv})$
\State $P \gets \text{GetCaliPara}(G',CaliSample)$
\State Begin \textbf{computing}
\While {$\sim$end of $InputVectors$}
\State $V_{in}[i] \gets InputVector[i]$
\State $I_{out}[i] = \text{CrossbarSim}(G', V_{in}[i])$
\State $Output[i] = \text{Calibration}(I_{out}[i],InputVector[i],P)$
\State $i \gets i + 1$
\EndWhile
\State $\text{Convolution result} \gets \text{reshape}(Output)$
\State End \textbf{computing} 
\end{algorithmic}
\label{alg_1}
\end{algorithm}

\begin{table}[t]
\caption{Explanations for functions in algorithm \ref{alg_1}}
\centering
\begin{tabular}{|l|p{6.5cm}|}
\hline
\textbf{Function} & \textbf{Explanation}\TBstrut\\ \hline
CrossbarSim       & Experiment verified crossbar simulator from \cite{hu_memristor-based_2018}\TBstrut\\ \hline
Conversion      & Solve $V_{conv} \cdot G = \mathrm{CrossbarSim}(V_{conv}, G')$ to get G'.\TBstrut \\ \hline
GetCaliPara      & Get 1st order poly fitting result $P$ by fitting crossbar\Tstrut \  output to ideal output of calibration samples.\Bstrut\\ \hline
Calibration      & Use $P$ and $InputVector[i]$ to map $I_{out}[i]$ to\Tstrut \  $Output[i]$. Here $InputVector[i]$ is needed, because in VMM $Y = XA$, if $A$ contains negative values, $Y$ can be calculated by $Y = X(A + c) - c*\mathrm{sum}(X)$, while c is a large enough scalar to shift $A$ to all positive.\Bstrut\\ \hline
\end{tabular}
\label{Table_functions}
\end{table}

\begin{figure}[t]
      \centering
      \includegraphics[width=0.4\textwidth]{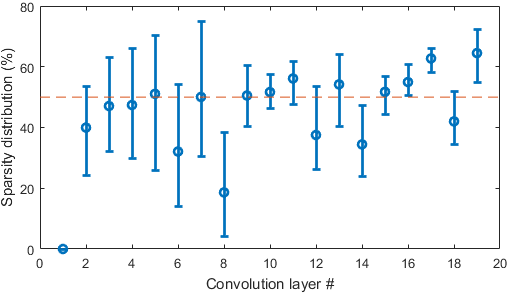}
      \caption{Input data sparsity at each convolution layer of ResNet-20}
      \label{fig_data_sparsity}
   \end{figure}
   
\begin{figure}[t]
      \centering
      \includegraphics[width=0.5\textwidth]{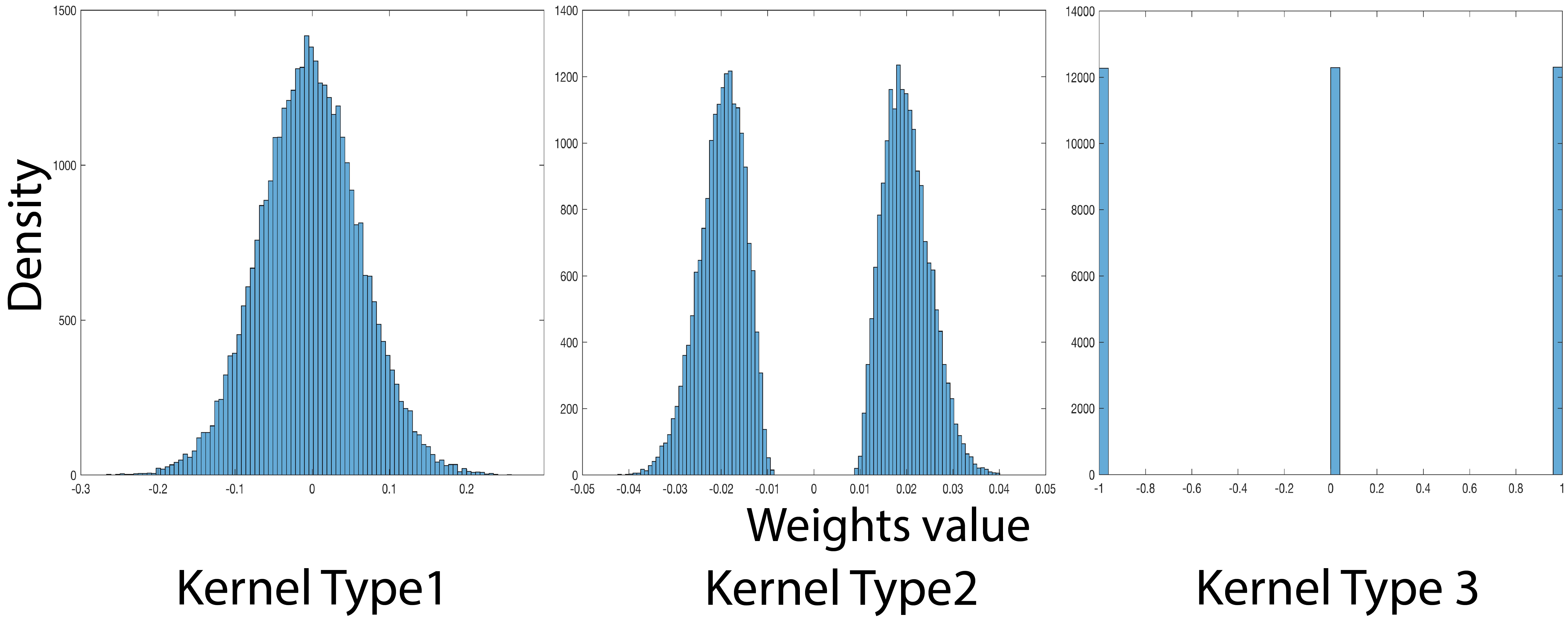}
      \caption{Three kernel types in CNN with different limitations on weight}
      \label{fig_kernel_type}
    \end{figure}
   
\subsubsection{\bf Data and kernel pattern in ResNet}
Data in ResNet has high sparsity due to ReLU. 
Fig. \ref{fig_data_sparsity} shows the data sparsity at each convolution layer of ResNet-20. 
The impact of data sparsity should be considered when choosing conversion signal as well as gathering calibration samples. 

similarly, we found that kernels in CNN have different distributions.
In Fig.\ref{fig_kernel_type} we list three typical kernel types regarding to their weight value distributions.
Usually there are three typical kernel types:
Kernel type 1 refers to a weight distribution close to Gaussian, usually it happens when training algorithms put no limitation on weight values, such as ResNet. Kernel type 2 refers to training algorithm that preventing weight values goes near zero\cite{han_deep_2015}. 
Kernel type 3 refers to Binary Neural Networks where weights can only be -1, 0(sparse), or 1\cite{courbariaux_binarized_2016}. 
It worth investigating how different kennel types in CNN impact the quality of crossbar-based convolution.   

\subsubsection{\bf Optimize conversion signal}
\begin{figure}[t]
      \centering
      \includegraphics[width=0.5\textwidth]{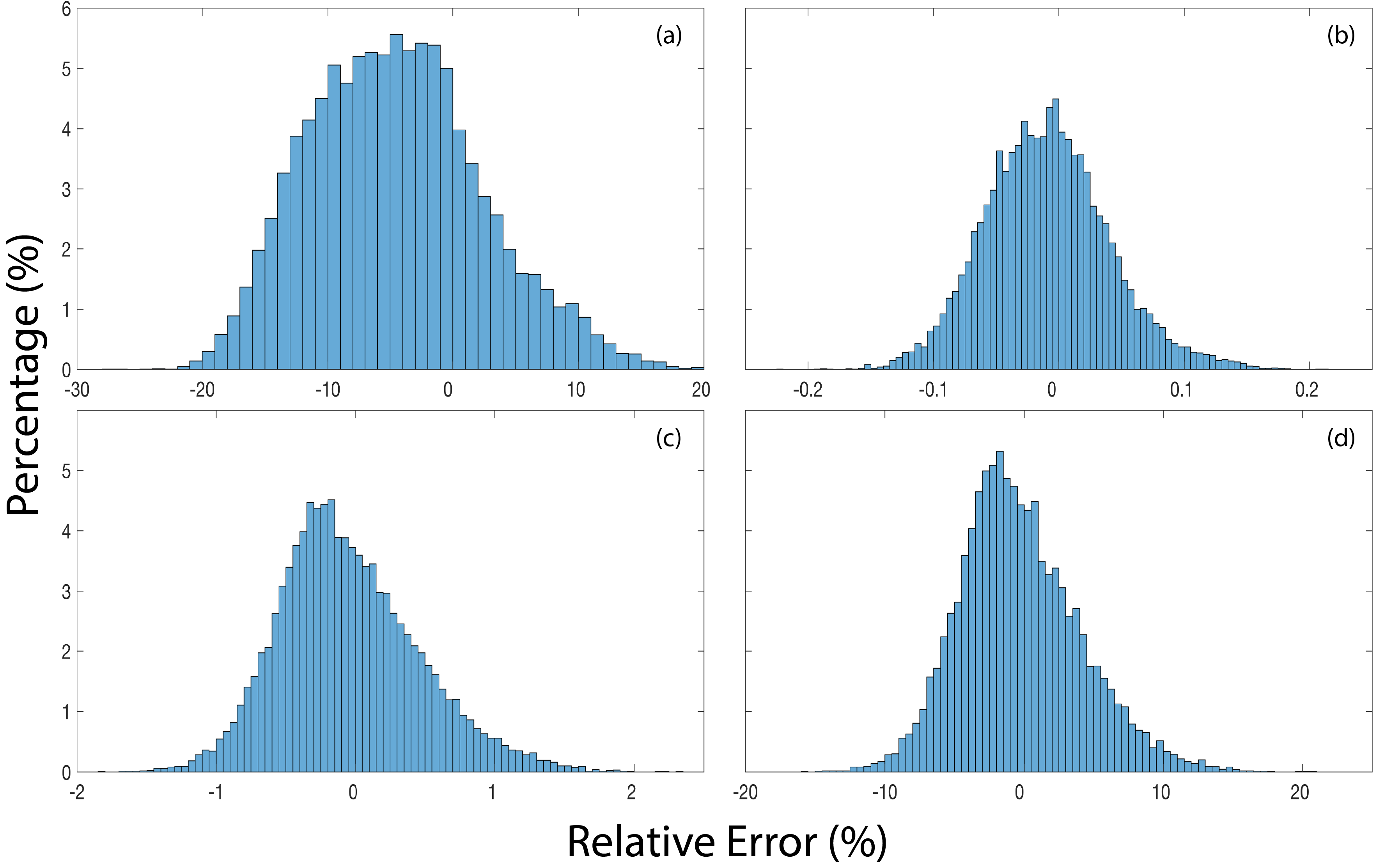}
      \caption{Relative error with different conversion signal amplitudes on crossbar(size: $144 \times 16$), (a),(b),(c),(d) are conversion signals with all 1, 0.1, 0.01, 0.001 respectively.}
      \label{fig_diffamp}
   \end{figure}

To better quantize the computing accuracy, we define relative error as below:
$$
\mathrm{Relative~Error} = \mathrm{Absolute~Error}/\mathrm{Output~Range} 
$$
While Absolute Error is the absolute error of crossbar output, and Output Range is the ideal convolution output range for each kernel.    
Relative error can be converted to output bit accuracy as below: 
$$
\mathrm{Bit~Accuracy} = log_2(1/\mathrm{Relative~Error} + 1)
$$
The original conversion algorithm takes the maximum input vector as its conversion signal, which works well with dense matrix and dense input signals. However in CNN, we need to consider the sparsity of data to optimize the conversion signal. By testing various different conversion signals across different kernels, we notice that the amplitude of conversion signal is critical, while the sparsity of conversion signal is not as important. Fig.\ref{fig_diffamp} shows the relative error distribution with different conversion signal amplitudes in crossbar with size $144 \times 16$. We found that a conversion signal with too large amplitude (all 1) will cause over compensation in the signal loss due to wire resistance, and a too small conversion signal (all 0.001) do have enough compensation and both of them result in large error in output.
So for crossbar size $144 \times 16$, all 0.1 signal appears to be the best conversion signal, and other conversion signal close to it generate similar error distribution.

\subsubsection{\bf Calibration}

\begin{figure}[t]
      \centering
      \includegraphics[width=0.5\textwidth]{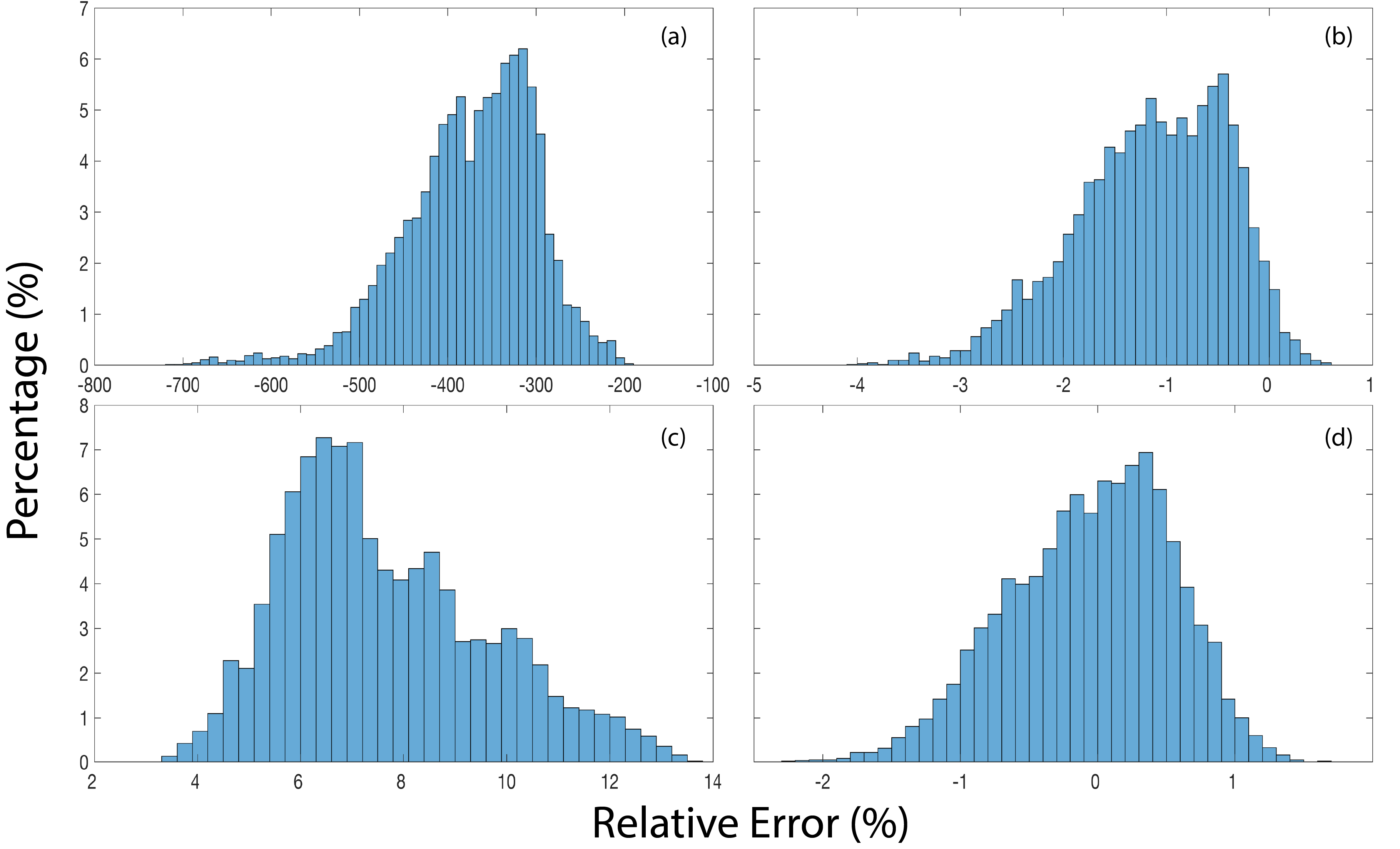}
      \caption{Calibration with different signal types on crossbar(size: $288 \times 32$), (a):Non-calibration, (b):random calibration signal, (c):random but decreased value(rand/cali\_iters),(d):chosen from shuffled input.}
      \label{fig_diff_signal_type}
   \end{figure}

In addition to original conversion algorithm, we add a calibration stage to further improve the result. It randomly picks 10 samples from the input data set, and runs a 1st order polynomial fit to fit crossbar output to ideal output. The generated fitting vector $P$ is fixed per crossbar, and can be easily embedded in ADC/DAC configurations.
Fig.\ref{fig_diff_signal_type} shows the relative error with different calibration signals, and here using shuffled input can achieve the best result. 


   






\section{\bf Result}

\subsection{\bf Simulation setup}

In our work, convolution and fully-connected (FC)layers are implemented by analog memristor crossbars with digital interface. Other functions, such as pooling, ReLU, batch normalization, etc., are processed by digital circuits. 
The CNN is offline-trained, then its kernels are converted to the conductance of memristors for inference.
Our crossbar simulator is configured as below: Lowest memristor resistance $R_{on}$ = 15k$\Omega$, Highest memristor resistance $R_{off}$ = 300k$\Omega$, wire resistance per segment is set to 1$\Omega$, sensing voltage = 0.2V. Input/Output resistance of crossbar are set to 1$\Omega$. Input voltage range is [0, 0.2V]. The crossbar simulation tool is adapted from \cite{hu_memristor-based_2018}. 


\subsection{\bf Individual convolution layer simulation}

\begin{figure}[tp]
      \centering
      \includegraphics[width=0.5\textwidth]{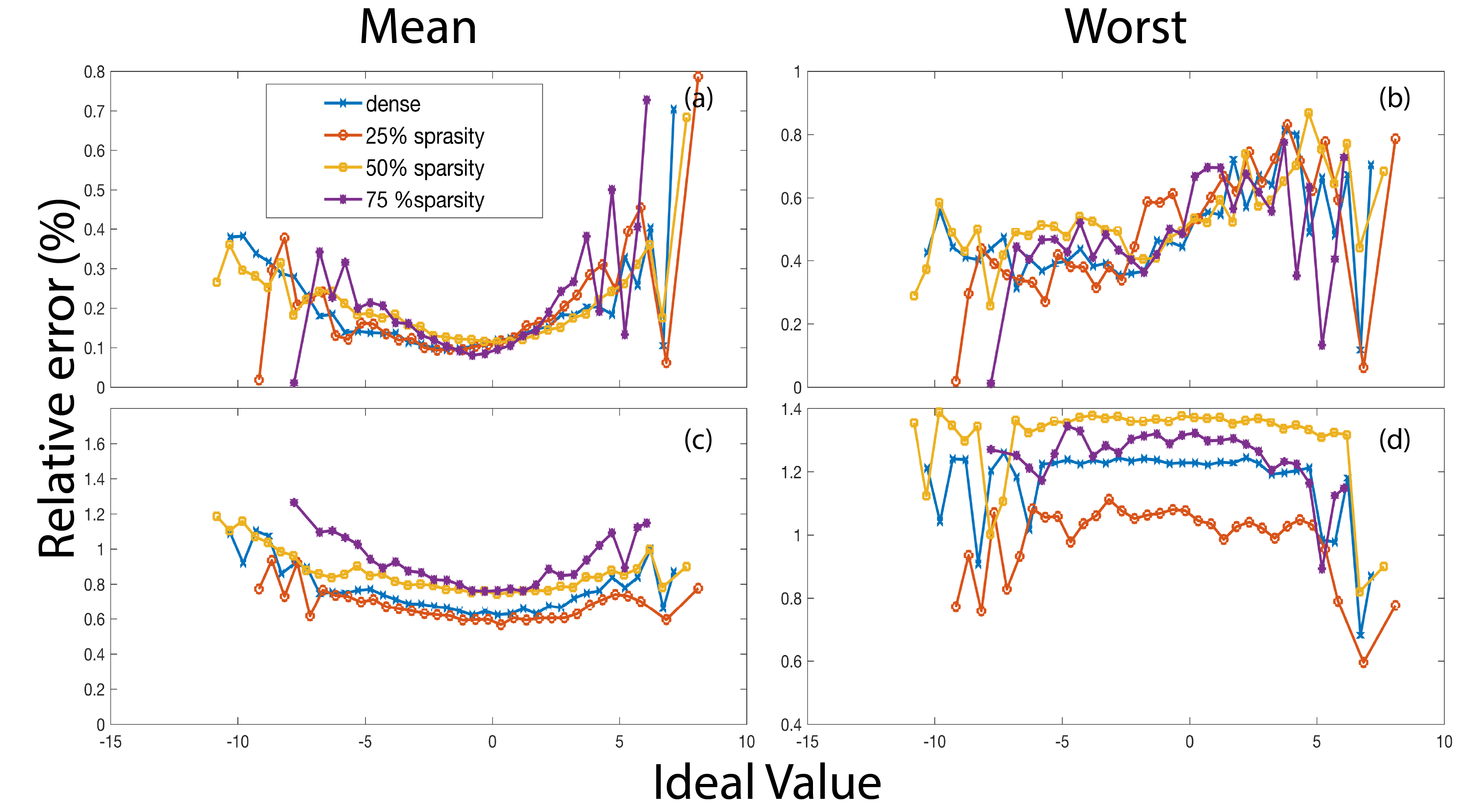}
      \caption{Relative error of $288 \times 32$ crossbar. (a),(b): improved conversion algorithm, (c),(d): original conversion algorithm}
      \label{fig_288x32}
   \end{figure}

\begin{figure}[t]
      \centering
      \includegraphics[width=0.4\textwidth]{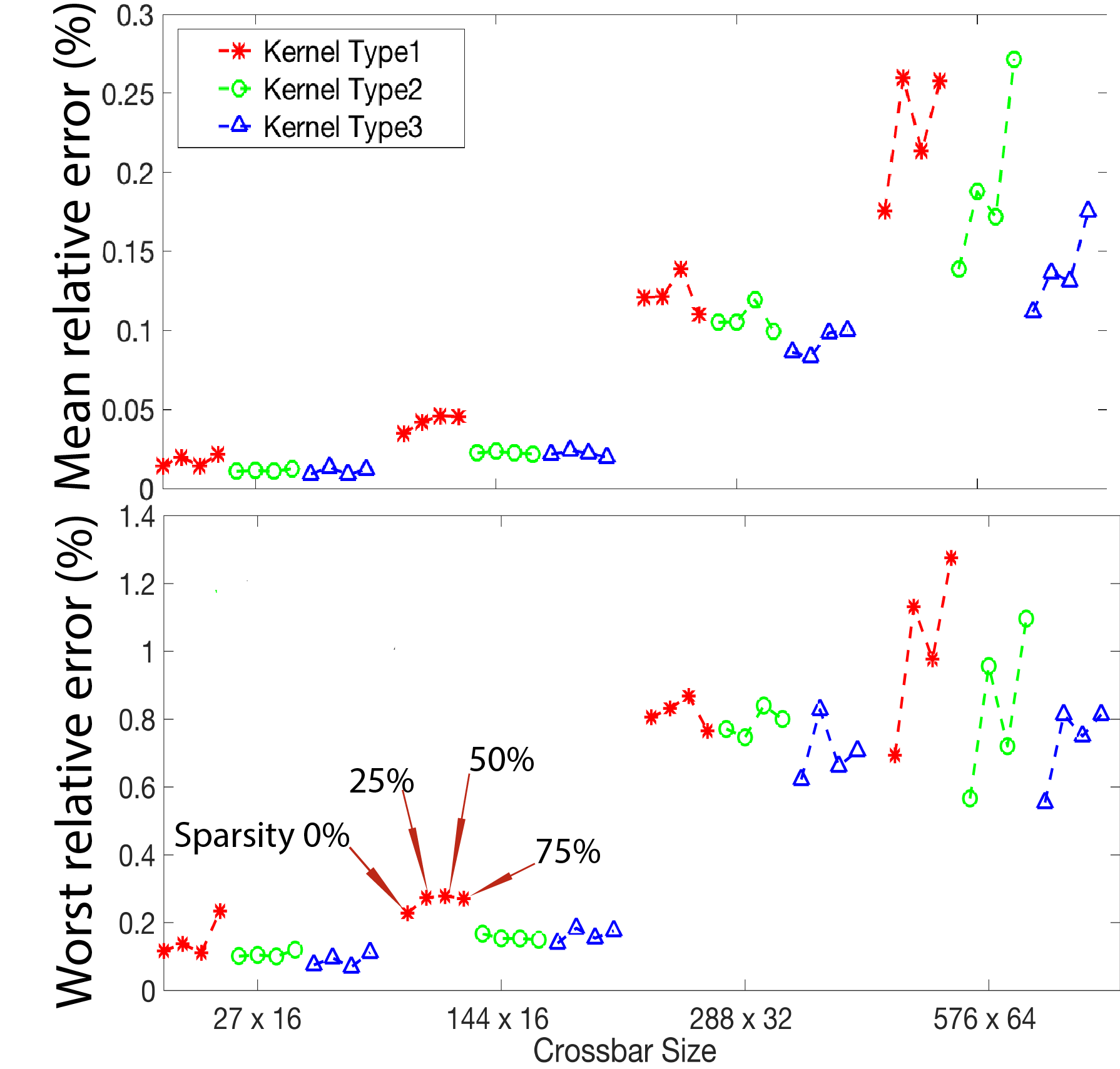}
      \caption{Mean/Worst relative error with different kernel types, data sparsities and crossbar sizes}
      \label{fig_kernel type}
   \end{figure}

We first run circuit simulation at individual convolution layers to understand the impact of input data sparsity, kernel type and size on convolution accuracy. 
Fig.\ref{fig_288x32} illustrate the impact of input data sparsity. 
There are three observations: first, our method provides $\sim$50\% better overall accuracy than the original conversion algorithm. Second, our method gives lower relative error when ideal value is small. Third, our method minimizes the impact of data sparsity comparing to original conversion algorithm.
Fig.\ref{fig_kernel type} summarizes the mean/worst relative error across three types of kernels with different input sparsity and crossbar sizes. It shows that our method is independent of kernel type and data sparsity. 


\subsection{\bf End-to-end simulation for ResNet}         

\begin{figure}[t]
      \centering
      \includegraphics[width=0.4\textwidth]{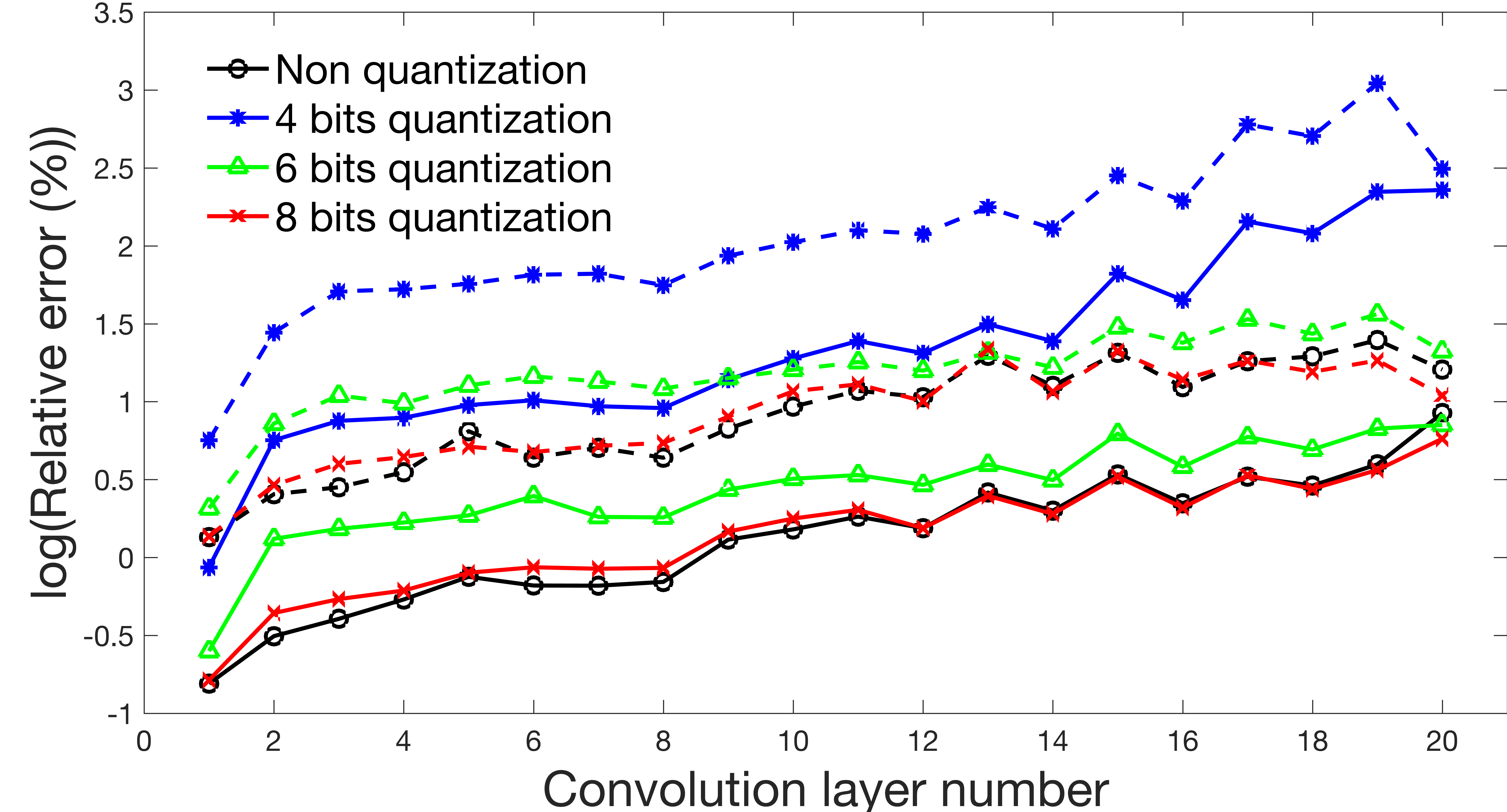}
      \caption{Error propagation along layers in ResNet-20, solid lines represent mean error, dash lines represent worst error.}
      \label{fig_layer_error}
   \end{figure}

\begin{figure}[t]
      \centering
      \includegraphics[width=0.4\textwidth]{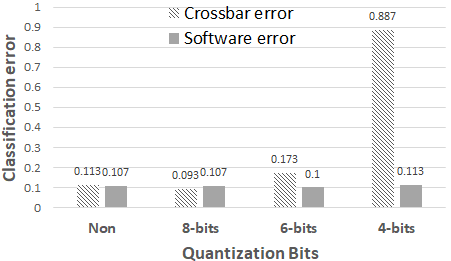}
      \caption{ResNet-20 result with different quantization settings}
      \label{fig_ResNet_result}
   \end{figure}

It is more important to predict how errors propagate and accumulate in deep neural networks, and evaluate its impact on the final classification result.
Fig.\ref{fig_layer_error} shows the error propagation at each convolution layer in ResNet-20. Fig.\ref{fig_ResNet_result} shows the classification result.
Different ADC/DAC quantization settings are applied and we can see that 8-bit quantization is necessary to prevent error accumulation. 
Without quantization, the error will propagate from beginning to the end and affect the classification result, which is also not practical for large-scale implementation. 
Due to long simulation time (15mins/image), we used a subset of cifar-10 containing 150 testing images. 
It appears that 8-bit quantization is a good balance between error suppression and information preservation, as it achieves even slightly better classification result than software.
6-bit or 4-bit quantization cause error accumulate through all layers and lead to a significant drop in classification accuracy. With 6-bit, error rate increased by 73\% from 10\% to 17.3\% , and 4-bit increases the error rate by a factor of 8 from 11.3\% to 88.7\%.  
   
\section{\bf Conclusions}

In this work, We take ResNet as case study to investigate how modern CNN performs on memristor-based implementations. 
Our methods includes an dense mapping and an improved conversion algorithm, which can achieve 0.25\% mean relative error ($\sim$ 8.6 bits) or 1.2\% worst relative error ($\sim$ 6.4 bits) for crossbar size $576 \times 64$.
We performed a rigorous analog simulation for every convolution layers to give an accurate prediction of error propagation in ResNet. We find that 8-bit ADC/DAC is necessary to prevent classification degradation.
Our method can be applied to general CNNs due to its independence of input data sparsity and kernel types.

\bibliographystyle{IEEE}
\bibliography{DPE.bib}  
\end{document}